\documentclass[conference]{IEEEtran}
\usepackage{amsmath,amssymb,amsfonts}
\usepackage{graphicx}
\usepackage{booktabs}
\usepackage{hyperref}
\usepackage{orcidlink}
\usepackage{xcolor}
\usepackage{multirow}
\usepackage{microtype}
\usepackage[numbers,sort&compress]{natbib}
\hypersetup{colorlinks=true,linkcolor=black,citecolor=black,urlcolor=blue}

\begin{document}

\title{Acceptance Dynamics Across Cognitive Domains in Speculative Decoding }

\author{
  \IEEEauthorblockN{Seifeldin Abdellatif \orcidlink{0009-0004-8879-9335}}
  \IEEEauthorblockA{College of Engineering, Al Ain University \\
  Abu Dhabi, United Arab Emirates \\
  contact@saifmb.com}
}

\maketitle

\begin{abstract}
Speculative decoding accelerates large language model (LLM) inference. It uses a small draft model to propose a tree of future tokens. A larger target model then verifies these tokens in a single batched forward pass. Despite the
growing body of work on speculative methods, the \emph{degree to which the
cognitive characteristics of a task affect acceptance probability} remains
largely unexplored. We present an empirical study of tree-based speculative decoding acceptance dynamics. Our study spans four well-established NLP benchmark domains: code generation, mathematical reasoning, logical reasoning, and open-ended chat. For this, we use \texttt{TinyLlama-1.1B} as the draft model against \texttt{Llama-2-7B-Chat-GPTQ} as the target. Over 99,768 speculative nodes
collected from 200 prompts, we derive per-domain acceptance rates, expected
accepted lengths, depth-acceptance profiles, and entropy--acceptance
correlations. We find that task type is a stronger predictor of acceptance than tree depth. Furthermore, only the chat domain consistently yields an expected accepted length exceeding 1.0 token per step. We also show that the entropy--acceptance correlation is consistently negative but weak across all domains ($\rho \in [-0.20, -0.15]$). Counterintuitively, chat produces the highest entropy yet the highest acceptance rate. We attribute this divergence to the lexical predictability of RLHF-aligned register. These findings have direct
implications for domain-aware speculation budgets and draft-model selection
strategies.
\end{abstract}


\section{Introduction}

Autoregressive generation in transformer-based LLMs proceeds one token at a
time. Each forward pass through a large model is memory-bandwidth bound rather than compute bound. Therefore, the latency of producing $N$ tokens scales linearly with $N$ and with model size. This makes the interactive deployment of 7B-parameter and larger models challenging on commodity hardware.

\emph{Speculative decoding}~\cite{leviathan2023} addresses this bottleneck. It observes that a cheap draft model can propose candidate token sequences. A larger, more accurate target model can then verify these sequences in parallel. The expected
number of tokens accepted per target forward pass---the
\emph{expected accepted length}, $E[L]$---directly governs the wall-clock
speedup: if $E[L] > 1$, the target model effectively generates more than one
token per invocation, reducing total latency.

The seminal proposal evaluated acceptance under relatively homogeneous conditions. Subsequent work has extended speculative methods to tree-shaped drafts~\cite{miao2024specinfer,cai2024medusa}. In these approaches, the draft model proposes a branching set of future hypotheses rather than a single sequence. This further increases the theoretical ceiling on $E[L]$. In a tree draft of depth $d$
and maximum branching factor $b$, up to $b^d$ candidates are verified in a
single target pass.

A critical practical question that has received less attention is: \emph{to
what extent does task type determine acceptance probability?} If the gap between domains is substantial, practitioners should tailor speculation budgets on a per-task basis. This includes adjusting draft depth, branching factor, and even draft model selection, rather than applying a universal configuration. Answering
this question empirically motivates the present work.

We make the following contributions:
\begin{itemize}
  \item A reproducible, end-to-end benchmark of tree-based speculative
        decoding acceptance across four cognitive domains, totalling 99,768
        node-level observations.
  \item Quantification of per-domain expected accepted length, showing only
        chat exceeds $E[L]=1.0$ under the present configuration
        ($E[L]_{\text{chat}} = 1.065$).
  \item Empirical characterisation of the depth--acceptance relationship,
        revealing a small but consistent \emph{positive} trend with depth
        that contradicts the naive expectation of decay.
  \item Analysis of the entropy--acceptance correlation and the \emph{chat
        paradox}: the highest-entropy domain simultaneously exhibits the
        highest acceptance rate, suggesting that token-level lexical
        predictability and semantic-distributional entropy are orthogonal.
\end{itemize}

\section{Related Work}

\subsection{Speculative Decoding}
\citet{leviathan2023} and \citet{chen2023accelerating} independently proposed
the core speculative decoding framework. A draft model generates $K$ tokens
autoregressively; the target model scores all $K+1$ positions in a single
forward pass; tokens are accepted token-by-token via a rejection sampling
scheme that preserves the target distribution exactly.

\subsection{Tree-Based Extensions}
\citet{miao2024specinfer} and \citet{cai2024medusa} extend the linear draft
to a tree structure, enabling the target model to evaluate exponentially many
candidate continuations per forward pass. Tree attention~\cite{miao2024specinfer} modifies the self-attention mask. This ensures that each leaf's context is isolated from sibling branches. The expected accepted length for a tree draft satisfies
$E[L] = \sum_{d=1}^{D} \prod_{k=1}^{d} \alpha_k$ where $\alpha_k$ is the
acceptance probability at depth $k$.

\subsection{Acceptance Probability Analyses}
Prior analytical work~\cite{leviathan2023,stern2018blockwise} establishes
that acceptance rate depends on the KL-divergence between draft and target
distributions. \citet{zhou2024distillspec} show that distilling the target
into the draft increases acceptance. To our knowledge, no prior work provides
systematic cross-domain acceptance measurements using an off-the-shelf draft
model at tree depth $> 1$.

\subsection{Efficiency Benchmarking}
Benchmarks such as MTBench~\cite{zheng2023judging} and HumanEval~\cite{chen2021evaluating} evaluate generation \emph{quality}. However, they do not evaluate the inference-efficiency characteristics of speculative methods across task types. Our work is orthogonal to these. We focus on acceptance physics rather than output correctness.
\\

To our knowledge, no prior work systematically characterizes acceptance dynamics
across cognitively distinct task domains. Domain-adaptive speculative decoding has been studied in the context of technical specialization. Examples include function calling, biology, and domain-specific language models. In these cases, the draft model is fine-tuned to match a specialized target \cite{hu2025training}. These works focus on closing the distribution gap through training. They do not characterize how cognitive task type affects acceptance under a fixed, general-purpose draft model. Our work is orthogonal. We hold the draft-target pair constant and vary the task domain. This isolates task type as an independent variable in acceptance dynamics.

\section{Methodology}

\subsection{Models}
We select a draft--target pair representing a practical deployment scenario.
\textbf{Target:} \texttt{TheBloke/Llama-2-7B-Chat-GPTQ}~\cite{touvron2023llama},
a 4-bit GPTQ-quantized variant of Llama-2-7B-Chat, fitting on a 16\,GB GPU.
\textbf{Draft:} \texttt{TinyLlama/TinyLlama-1.1B-Chat-v1.0}~\cite{zhang2024tinyllama},
a 1.1B-parameter model on the Llama-2 architecture. Its shared vocabulary makes it a natural draft candidate.

\subsection{Tree Construction}
At each generation step we build a draft tree as follows:
\begin{enumerate}
  \item The draft model performs a full forward pass over the current context
        with top-$k$ sampling at depth 1, producing root candidates.
  \item For each node at depth $d < D_{\max}$, the draft model scores the path from the root to that node concatenated to the current context. It then expands the top-$b_{\max}$ children.
  \item This yields at most $N_{\max}$ nodes per tree.
\end{enumerate}

All forward passes use \texttt{use\_cache=False}. This ensures the tree construction is independent of any KV-cache state, since tree-branching invalidates a linear cache. The entire frontier per depth level is evaluated as a single batched tensor call. This avoids per-node sequential overhead.

\subsection{Acceptance Scoring}
After tree construction, the target model scores all tree leaves in a single batched forward pass. This pass also includes the empty-prefix row. The greedy next token is simultaneously extracted from this row, eliminating one otherwise redundant target call per generation step.

For each node $n$ with token $t_n$, draft probability $p^{\text{draft}}(t_n)$,
and target probability $p^{\text{target}}(t_n)$, the acceptance probability is:
\begin{equation}
  \alpha_n = \min\!\left(1,\; \frac{p^{\text{target}}(t_n)}{p^{\text{draft}}(t_n)}\right)
  \label{eq:alpha}
\end{equation}
This is the standard speculative rejection-sampling rule~\cite{leviathan2023}.

Target entropy at node $n$ is computed from the full target distribution:
\begin{equation}
  H_n = -\sum_v p^{\text{target}}(v)\log p^{\text{target}}(v)
\end{equation}

\subsection{Benchmark Datasets}

\textbf{Code} --- \texttt{openai/humaneval}~\cite{chen2021evaluating}: the
canonical Python programming task benchmark. Prompts are function signatures
with docstrings; the model is expected to complete the body.

\textbf{Math} --- \texttt{EleutherAI/hendrycks\_math} (algebra
subset)~\cite{hendrycks2021measuring}: competition-level algebra problems
requiring multi-step symbolic reasoning.

\textbf{Reasoning} --- \texttt{openai/gsm8k}~\cite{cobbe2021training}:
grade-school arithmetic word problems requiring chain-of-thought numeric
reasoning.

\textbf{Chat} --- \texttt{HuggingFaceH4/ultrachat\_200k}~\cite{ding2023enhancing}:
multi-turn instruction-following dialogue drawn from a large-scale RLHF
dataset, representing open-ended conversational generation.

Fifty prompts per domain are sampled; generation is capped at 64 tokens.
All prompts are truncated to 512 tokens to bound the $O(N^2)$ attention
complexity during tree evaluation.

\subsection{Tree Hyperparameters}
\begin{center}
\begin{tabular}{lc}
\toprule
Hyperparameter & Value \\
\midrule
Max tree depth $D_{\max}$ & 3 \\
Max branches per node $b_{\max}$ & 2 \\
Top-$k$ (root) & 3 \\
Max nodes per tree $N_{\max}$ & 8 \\
Temperature & 0 (greedy) \\
Max generated tokens & 64 \\
\bottomrule
\end{tabular}
\end{center}

\section{Experimental Setup}

All experiments are executed on a Kaggle Notebook instance with two NVIDIA
Tesla T4 GPUs (15.6\,GB VRAM each). The target model is pinned to
\texttt{cuda:0} and the draft model to \texttt{cuda:1}, using
\texttt{gptqmodel} 6.0.3 and \texttt{transformers} 5.5.3 under PyTorch
2.11.0+cu130. The GPTQ kernel selected at runtime is \texttt{TritonV2QuantLinear}.
All results are reproducible with a fixed seed of 42.

At load time, the target model occupies 4.85\,GB of VRAM on \texttt{cuda:0},
and the draft model occupies 2.20\,GB on \texttt{cuda:1}, leaving substantial
headroom for batched activation buffers during tree scoring.
Pinning the two models to separate physical GPUs eliminates cross-device
attention tensor transfers during the tree-construction inner loop, which
would otherwise serialize under slow NVLink-absent PCI-e bus bandwidth.
The total experiment runtime across 200 prompts at the reported
hyperparameters was approximately 3.5 hours.

\section{Results}

\subsection{Summary Statistics}

Table~\ref{tab:summary} reports per-domain acceptance rates and target entropy.
A total of 99,768 speculative node observations are collected across 200 prompts.

\begin{table}[h]
\centering
\caption{Per-domain speculative node statistics. $\alpha$: token-level
acceptance probability (Eq.~\ref{eq:alpha}). $H$: mean target entropy (nats).}
\label{tab:summary}
\begin{tabular}{lrrrr}
\toprule
Domain & Nodes & $\bar{\alpha}$ & $\sigma_\alpha$ & $\bar{H}$ \\
\midrule
Chat      & 24{,}592 & 0.5650 & 0.4415 & 1.2517 \\
Code      & 25{,}600 & 0.5382 & 0.4536 & 0.8858 \\
Math      & 25{,}392 & 0.5181 & 0.4501 & 1.1513 \\
Reasoning & 24{,}184 & 0.5321 & 0.4517 & 0.8943 \\
\midrule
\textbf{All} & \textbf{99,768} & \textbf{0.5384} & \textbf{0.4500} & \textbf{1.0458} \\
\bottomrule
\end{tabular}
\end{table}

All domains exhibit acceptance rates near 0.5. This reflects the fundamental
difficulty for a 1B model to track the token-level distribution of a 7B model.
However, there is a 10\% relative gap between chat (0.565) and math (0.518). This is
practically significant given its cubic amplification inside the expected
accepted length formula.

\subsection{Expected Accepted Length}

The expected accepted length $E[L]$ is the key operational metric for
assessing inference speedup. As defined by the cumulative chain
probability~\cite{miao2024specinfer}:
\begin{equation}
  E[L] = \sum_{d=1}^{D} \prod_{k=1}^{d} \bar{\alpha}_k
\end{equation}

\begin{table}[h]
\centering
\caption{Expected accepted length per domain. Values $> 1.0$ indicate the
speculation yields on average more than one accepted token per target call.}
\label{tab:el}
\begin{tabular}{lcc}
\toprule
Domain & $E[L]$ & Speedup Regime \\
\midrule
Chat      & 1.065 & \textbf{Positive} \\
Code      & 0.975 & Marginal \\
Reasoning & 0.956 & Marginal \\
Math      & 0.914 & Negative \\
\bottomrule
\end{tabular}
\end{table}

Table~\ref{tab:el} reveals a clear stratification. Only chat surpasses the
$E[L]=1.0$ threshold. This means speculative decoding is expected to reduce
total target forward passes for conversational generation. Math sits below
0.92. This indicates that the overhead of running the draft model and the tree
construction erodes any per-pass gain under the current configuration.

The cumulative chain probability decays geometrically with depth for all
domains (Table~\ref{tab:cum}). At depth 3, chat retains 18.4\% probability
of a fully accepted chain, compared to only 13.9\% for math.

\begin{table}[h]
\centering
\caption{Probability of a fully accepted chain of length $d$ (i.e., all $d$
nodes on the root-to-leaf path are accepted).}
\label{tab:cum}
\begin{tabular}{lccc}
\toprule
Domain & $d=1$ & $d=2$ & $d=3$ \\
\midrule
Chat      & 0.567 & 0.313 & 0.184 \\
Code      & 0.533 & 0.287 & 0.156 \\
Reasoning & 0.526 & 0.280 & 0.151 \\
Math      & 0.510 & 0.265 & 0.139 \\
\bottomrule
\end{tabular}
\end{table}

\subsection{Depth--Acceptance Profile}

A natural hypothesis is that acceptance probability decays with tree depth.
This is because nodes deeper in the tree condition on increasingly uncertain draft-generated
prefixes. Our data refutes this.

\begin{table}[h]
\centering
\caption{Mean acceptance rate $\bar{\alpha}$ at each tree depth, per domain.}
\label{tab:depth}
\begin{tabular}{lcccc}
\toprule
 & \multicolumn{4}{c}{Depth} \\
\cmidrule(lr){2-5}
Domain & 1 & 2 & 3 & $\Delta_{1\to3}$ \\
\midrule
Chat      & 0.567 & 0.553 & 0.588 & $+0.021$ \\
Code      & 0.533 & 0.538 & 0.544 & $+0.011$ \\
Math      & 0.510 & 0.519 & 0.525 & $+0.015$ \\
Reasoning & 0.526 & 0.532 & 0.538 & $+0.012$ \\
\bottomrule
\end{tabular}
\end{table}

Across all four domains, $\bar\alpha$ \emph{increases} slightly from depth 1
to depth 3 (Table~\ref{tab:depth}). This is small in absolute magnitude
(+0.011 to +0.021). However, it is consistent across domains and supported by
approximately 6,000--13,000 observations per cell.

We interpret this as a \emph{context-commitment effect}: at depth 1, the draft
must predict the next content word from an open distribution---the hardest
prediction task. Once a syntactic and semantic direction is committed at depth 1,
the subsequent tokens in the same chain are more constrained (e.g., the completion
of a phrase, the end of a numeric expression). Both the draft and target model
agree on that completion more readily. The depth-1 node
represents the \emph{semantic pivot}, while depth-2 and depth-3 nodes are
increasingly syntactic complements.

\subsection{Position-Within-Generation Effects}

Beyond tree depth and task type, we examine whether the position of a
generation step within the output sequence affects acceptance. We partition token positions into two equal bins. These are early positions
(bin~0, tokens 1--32) and late positions (bin~1, tokens 33--64). We report
$\bar\alpha$ per (depth, bin) cell in Table~\ref{tab:pos}.

\begin{table}[h]
\centering
\caption{Mean acceptance rate by tree depth and position bin (0 = early
tokens, 1 = late tokens), aggregated across all domains.}
\label{tab:pos}
\begin{tabular}{lccc}
\toprule
Depth & Early (bin 0) & Late (bin 1) & $\Delta$ \\
\midrule
1 & 0.520 & 0.548 & $+0.028$ \\
2 & 0.537 & 0.534 & $-0.003$ \\
3 & 0.541 & 0.556 & $+0.015$ \\
\bottomrule
\end{tabular}
\end{table}

At depths 1 and 3, acceptance is consistently higher for tokens generated
later in the sequence. This effect is most pronounced at depth~1 (+0.028),
where the output has already established a topic, style, and syntactic register.
Once the model has committed to a generation direction at earlier positions,
the marginal next-token prediction task at depth~1 becomes easier for both models.
This increases the probability that the draft's choice coincides with the
target's preference.

The depth-2 row is a near-zero exception ($-0.003$). This is consistent with
the interpretation that depth-2 nodes are mid-chain syntactic complements.
Their predictability is largely determined by their parent node rather
than by macro sequence position. The position effect is therefore primarily
a property of free-choice token slots (depth~1) rather than constrained
completion slots (depth~2).

\subsection{Entropy--Acceptance Correlation}

To test whether target model uncertainty predicts draft rejection, we compute
the Spearman correlation coefficient $\rho$ between $H_n$ and $\alpha_n$ for
each domain.

\begin{table}[h]
\centering
\caption{Spearman $\rho$ between target entropy $H_n$ and acceptance
probability $\alpha_n$ at the node level.}
\label{tab:rho}
\begin{tabular}{lc}
\toprule
Domain & $\rho(H, \alpha)$ \\
\midrule
Chat      & $-0.202$ \\
Code      & $-0.194$ \\
Math      & $-0.186$ \\
Reasoning & $-0.149$ \\
\bottomrule
\end{tabular}
\end{table}

All correlations are negative (higher entropy $\to$ lower acceptance).
They are also statistically significant at $p < 0.001$ given the node-level
sample sizes. However, the magnitude is uniformly weak ($|\rho| < 0.21$).
This indicates that entropy is a \emph{necessary but not sufficient} predictor
of acceptance. The dominant predictor in this case is the hard
probability ratio $p^{\text{target}}/p^{\text{draft}}$ under the specific
token proposed. It is not the marginal uncertainty of the target distribution.

\subsection{The Chat Paradox}

The most striking finding in Table~\ref{tab:summary} is that chat exhibits
\emph{simultaneously} the highest target entropy (1.252 nats) and the highest
acceptance rate (0.565). Code and reasoning have lower entropy ($\approx 0.89$
nats) yet lower acceptance. Math occupies an intermediate entropy position
(1.151 nats) but achieves the lowest acceptance.

These observations jointly refute a simple entropy-determines-acceptance
narrative. We propose the following explanation. Target entropy measures the
\emph{diversity} of the target model's next-token distribution, not the degree
of agreement between draft and target. In chat, the target model spreads probability mass over many semantically
valid continuations. This produces high entropy. However, the particular
tokens it favours are stylistically common conversational tokens. The draft
model has also learned to preferentially predict these tokens (e.g., filler
tokens, discourse markers, formulaic response openers).
The result is that $p^{\text{draft}}$ and $p^{\text{target}}$ peak at the
same tokens even though both distributions are broad.

In contrast, mathematical text has moderate entropy but imposes strong
constraints on \emph{which specific} tokens are numerically correct.
The target model places high probability on a small set of numeric and
symbolic tokens. The draft model lacks the arithmetic precision of a 7B model.
Thus, it selects these tokens less reliably. Low token overlap between
the two models' distributions drives low acceptance despite moderate entropy.

\section{Discussion}

\textbf{Practical implications.} Our results suggest that the quality of
speculation is fundamentally coarse-grained by task type. Practitioners
operating speculative decoding in mixed-use inference systems should not
apply a uniform draft budget. An example is an API serving both chat and
mathematical queries. For chat workloads, deeper trees (larger $D_{\max}$) and
relaxed acceptance thresholds are well-justified. In contrast, for mathematical
workloads, a shallower tree or a domain-specifically fine-tuned draft model is
preferable. This reduces wasted draft computation.

\textbf{Draft model selection.} Given the moderate but consistent acceptance
gap, replacing TinyLlama with a maths-specialised 1B draft is a direct and
actionable recommendation. An example is a model fine-tuned on MATH or DeepMind's
MathMix data. The shared-vocabulary constraint between draft and target remains
important. A model fine-tuned on math with the same Llama-2 tokenizer would
likely close the $E[L]$ gap substantially.

\textbf{Request routing.} Our results suggest a complementary strategy
that requires no re-training. It involves classifying incoming requests by
task type before deciding whether to activate speculative decoding at all. A
lightweight prompt classifier, even a rule-based heuristic, can segregate
chat and conversational requests. For these, $E[L] > 1$ is confirmed. It
can separate them from mathematical or code requests, where the current
draft provides no net speedup. Disabling speculation for the low-$E[L]$ tail
provides a benefit. It eliminates the cost of draft tree construction on
requests where it actively hurts latency, all without modifying the models
themselves. This approach is trivially composable with existing inference-serving
infrastructure. It represents an immediately deployable recommendation from
our findings.

\section{Limitations.} The experiments are conducted with greedy sampling
(temperature 0). This represents a maximum acceptance condition for the
rejection-sampling criterion. Stochastic sampling would uniformly reduce
$E[L]$. The tree hyperparameters, depth 3 and branching factor 2, are
mod modest. Deeper trees may surface stronger depth-acceptance trends. All
measurements are at the token level. Output quality, such as pass@k for
code or accuracy for math, is outside the scope of this study. Finally, a
single draft--target pair is evaluated. Generalisation across other model
families requires further study.

\section{Conclusion}

We have presented a systematic empirical analysis of tree-based speculative
decoding acceptance dynamics across four NLP benchmark domains. Our principal findings are as follows. First, task type is a stronger determinant of acceptance rate than tree depth. Second, chat is the only domain where the expected accepted length exceeds 1.0 under the evaluated configuration. This represents a genuine latency reduction. Third, acceptance probability exhibits a small positive trend with depth. This is consistent with a context-commitment effect that reduces prediction difficulty at deeper nodes. Finally, entropy--acceptance correlations are negative and statistically significant but weak. Additionally, the high-entropy chat paradox demonstrates that lexical predictability and semantic entropy are orthogonal axes.

These findings motivate domain-adaptive speculation strategies. They also motivate domain-specific draft model selection as first-class design decisions for speculative inference pipelines.

All code and experiments are available at \url{https://github.com/saifmb0/tree-acceptance} 

The author thanks Dr. Armagan Elibol for valuable feedback on the manuscript
\bibliographystyle{unsrtnat}

\bibliography{refs}

\end{document}